# Data imputation and comparison of custom ensemble models with existing libraries like XGBoost, Scikit learn, etc. for Predictive Equipment failure

Tejas Y. Deo

Department of Mechanical Engineering, PES's College of Engineering Pune, Savitribai Phule Pune University, Pune

Email ID: tejasdeo10@gmail.com

**Abstract:**

This paper presents comparison of a custom ensemble models with the models trained using existing libraries Like Xgboost, Scikit Learn, etc. in case of predictive equipment failure for the case of oil extracting equipment setup. The dataset that is used contains many missing values and the paper proposes different model-based data imputation strategies to impute the missing values. The architecture and the training and testing process of the custom ensemble models are explained in detail.

**Keywords:** Predictive Equipment Failure, Oil extracting equipment, Xgboost, Scikit Learn, Custom ensemble models, Artificial intelligence

## 1. Introduction:

Machines play a very important role in our day-to-day life, from using a toaster every morning to a car to commuting every day. Now if one of them started failing every day, it would have a huge impact on us. Humans invented machines to make their life easier and now, we are more dependent on them. The quality of the machine is not just dependent on how useful and efficient it is, but also on how reliable it is. Further along with reliability comes maintenance [1,2]. So, even if the quality of machine is excellent and is extremely useful, it won't sum up unless and until we have the reliability factor coming into the picture. This reliability can be maintained by knowing the machine insights and repairing it or replacing and maintaining it at the right time. As it is rightly said that, "*prevention is better than cure*", similarly for machines we need to perform the repairing or the maintenance work at the right time before the machine/equipment fails [3,4]. Fig.1 depicts evolution of maintenance strategies. Real-time health monitoring of machining equipment's in the industry is of utmost importance and it is a classic example of Industry 4.0. Health or condition monitoring of machining equipment's can be twofold: detecting/predicting the type of failure and estimating the remaining useful life of the equipment [5,6]. This paper focuses on predicting the type of failure occurring on the machining equipment. Predictive equipment failure is basically used to predict/anticipate the upcoming failures in machines based on the data pattern. By predicting the upcoming failures, one can stop the machinery and prevent the unnecessary failures, save maintenance cost, reduce the unwanted downtime, and maintain the quality of the workpiece [7,8]. The paper presents comparison of custom ensemble models with the models trained using existing libraries Like Xgboost, Scikit Learn, etc. in case of predictive equipment failure for the case of oil extracting equipment setup. The dataset that is used contains many missing values and the paper proposes different model-based data imputation strategies to impute the missing values. The architecture and the training and testing process of the custom ensemble models are explained in detail.

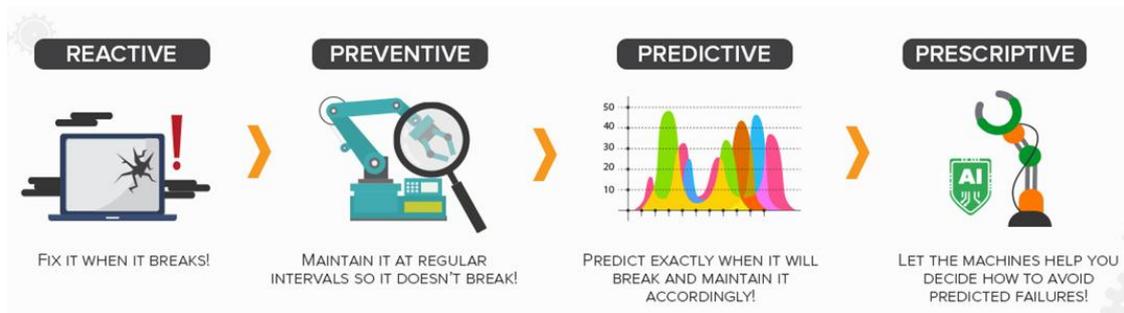

Fig.1 Evolution of Maintenance Strategies

## 2. Data Description and Source

The data is of ConocoPhillips Company and is taken from Kaggle. This company is mainly responsible for hydrocarbon production. And so, it also extracts oil from oil wells. The data which they have given is of oil wells (stripper wells to be specific).

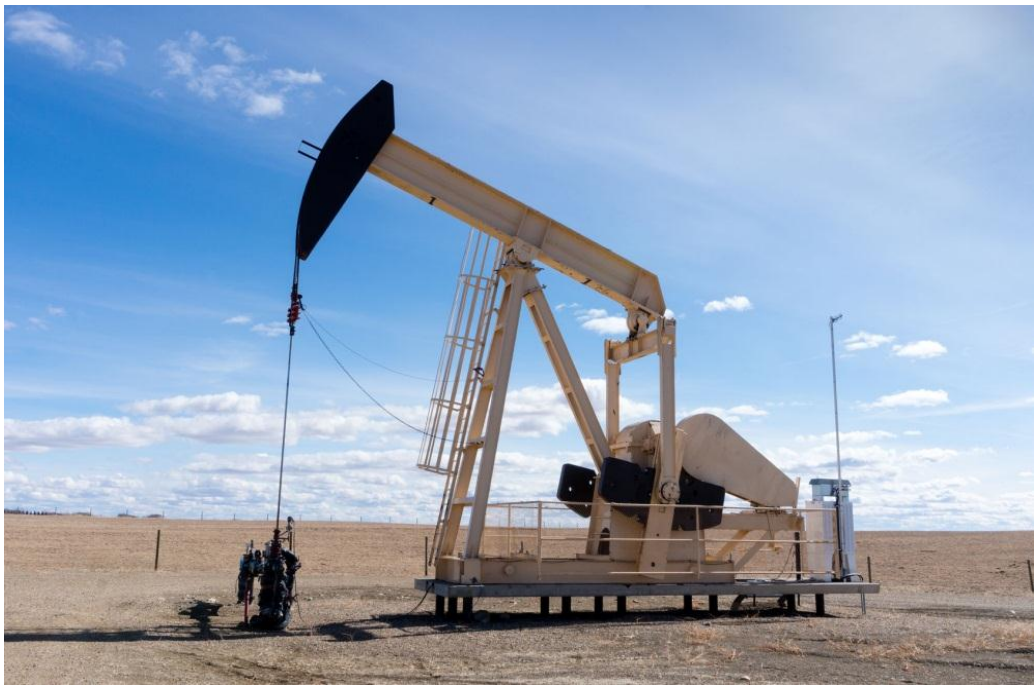

Fig.2 Oil extracting equipment setup [9]

Stripper well or marginal well is an oil or gas well that is nearing the end of its useful life. In simple words, the wells are not able to produce the expected amount of oil per day. 85% of the wells in the United States are stripper wells now. These stripper wells are responsible for significant amount of oil production. The data set provided has documented failure events that occurred on, on-surface equipment and down-hole equipment. For each failure event, data has been collected from over 107 sensors (attached on the mechanical equipment's) that collect a variety of physical information from the surface (on-surface equipment) and below the ground (downhole equipment). Fig. 2 show how the have mechanical equipment is usually placed on the surface i.e., on the ground and also below the ground to extract oil.

**3. Business Requirements**

In this section business requirements are discussed with the help of business problem, usage and constrains.

*3.1 Business Problem*

Stripper wells are attractive for the company due to their low capital intensity and low operational costs. Due to these reasons, the profit margin given by these stripper wells is comparatively large. This cash is then used by the company to fund operations that require more money. For example, the ConocoPhillips company uses the funds from the West Texas Conventional operations (it is named as conventional operation because, the main export business in Texas is Petroleum, Coal Products and oil) which serves as a cash flow to fund more expensive projects in the Delaware basin and other unconventional plays across the United States. So, the business problem is to maintain this steady cash flow from the stripper wells and make sure that the maintenance cost of these stripper wells is as less as possible, to increase the profit margin and funds for the rest of the unconventional projects in the United States. But, as with all mechanical equipment's, things break and when things break money is lost in the form of repairs and lost oil production. It is necessary to prevent this failure in order to maintain a steady cash flow from the stripper wells to other operations.

*3.2 Business Usage*

- Prevent the failure of equipment's
- Reduce maintenance/repairing cost
- Cut unwanted downtime

*3.3. Business Constraints*

- No low latency Constraints: This is because the goal is to predict the upcoming failure and not predict the real time failure. So, it is possible to use complex models like ensemble models.
- High Precision and Recall: High Precision because, the aim is to perfectly predicts the surface or downhole failure. This is a 2-class classification problem which demonstrates the types of failure. It should not happen that the true/actual failure was for on-surface equipment and the crew sent a workover rig to fix the downhole equipment as the model predicted it to be a downhole failure. Now this would increase the unwanted downtime as the crew is looking for the failure at the wrong place. High Recall because, it is necessary to predict all the failures and not miss a single one.

Missing even a single failure would increase the maintenance cost i.e., the repairing cost, increase the unwanted downtime and loss in the form of oil and cash.

**4. Existing Approaches & Application of Machine Learning**

The job is to predict the type of failures (on-surface failure and down-hole/downhole failure). This information can be used to send crews to the well location to fix the equipment on the surface or, send a workover rig to the well to pull the down-hole equipment (equipment present below the ground level which is extracting the oil) and address the failure. Predicting the equipment failure on the surface and below the ground is what is required. The dataset is highly imbalanced where there are more data points for ON-SURFACE failure i.e., Class 0 and less data points for DOWNHOLE failure i.e., Class-1. To give equal importance to both the class labels, the Key parameter indicator (KPI) is "Macro F1 Score".

The dataset provided has 2 types of data, the "MEASURES DATA" (which store a single measurement from the sensor) and, the "HISTOGRAM DATA" (they have data recorded from the sensors for 10-time steps). Also, this dataset has large number of missing values i.e., "NAN values". The general overview of the existing approaches is that, most people imputed the missing values with either 0, mean or median. Some have analysed the data and have come to a conclusion that the "Histogram data" is not useful and so they just discarded it. Others have done feature engineering and have come up with new features from the 'Measures data'. According to the Business Constraints, there are no low latency requirements and so, almost all of them have used existing ensemble models like Random Forest, XGBoost, AdaBoost, CATBoost, etc and have got good results.

**5. A Novel Approach for EDA (Exploratory Data Analysis) and Data Pre-Processing**

From Fig. 3, it is evident that the dataset is highly imbalanced and has more data points for Surface Failure and less data points for Downhole Failure.

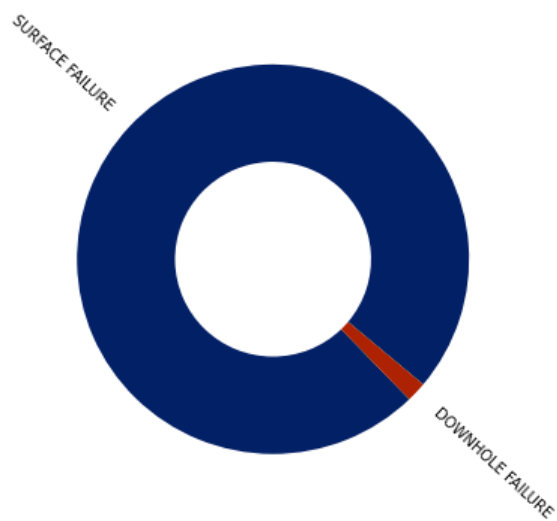

Fig.3 Data Imbalance Overview

Surface failure is 98.33% (blue portion) and downhole failure is 1.67% (orange portion). First, to separate the Measures columns (which make the Measures dataset) and the Histogram columns (which makes the Histogram dataset) from the input data. Further it is analysed and pre-processed separately. The analysis and pre-processing are done on the training data. Equivalent pre-processing will be done on the test data using the pre-trained data pre-processing models.

|   | sensor1_measure | sensor3_measure | sensor4_measure | sensor5_measure | sensor6_measure | sensor8_measure | sensor9_ |
|---|---|---|---|---|---|---|---|
| 0 | 76698.0 | 2.130706e+09 | 280.0 | 0.0 | 0.0 | 2551696.0 | |
| 1 | 33058.0 | 0.000000e+00 | NaN | 0.0 | 0.0 | 1393352.0 | |
| 2 | 41040.0 | 2.280000e+02 | 100.0 | 0.0 | 0.0 | 1234132.0 | |
| 3 | 12.0 | 7.000000e+01 | 66.0 | 0.0 | 10.0 | 2668.0 | |
| 4 | 60874.0 | 1.368000e+03 | 458.0 | 0.0 | 0.0 | 1974038.0 | |

5 rows × 94 columns

Fig.4 Measures Data Sample

|   | sensor7_histogram_bin0 | sensor7_histogram_bin1 | sensor7_histogram_bin2 | sensor7_histogram_bin3 | sensor7_histogram_ |
|---|---|---|---|---|---|
| 0 | 0.0 | 0.0 | 0.0 | 0.0 | 372 |
| 1 | 0.0 | 0.0 | 0.0 | 0.0 | 182 |
| 2 | 0.0 | 0.0 | 0.0 | 0.0 | 16 |
| 3 | 0.0 | 0.0 | 0.0 | 318.0 | 22 |
| 4 | 0.0 | 0.0 | 0.0 | 0.0 | 437 |

5 rows × 70 columns

Fig.5 Histogram Data Sample

*5.1 Analysis and pre-processing of the measures data (shape = (60000, 94))*

In Fig.6, the "describe()" function shows that most of the columns have a value close to 0 or have large values.

```
1  measures_df.describe()
```

|   | sensor1_measure | sensor3_measure | sensor4_measure | sensor5_measure | sensor6_measure | sensor8_measure |
|---|---|---|---|---|---|---|
| count | 6.000000e+04 | 5.666500e+04 | 4.513900e+04 | 57500.000000 | 57500.000000 | 5.935500e+04 |
| mean | 5.933650e+04 | 3.560143e+08 | 1.906206e+05 | 6.819130 | 11.006817 | 1.809931e+06 |
| std | 1.454301e+05 | 7.948749e+08 | 4.040441e+07 | 161.543373 | 209.792592 | 4.185740e+06 |
| min | 0.000000e+00 | 0.000000e+00 | 0.000000e+00 | 0.000000 | 0.000000 | 0.000000e+00 |
| 25% | 8.340000e+02 | 1.600000e+01 | 2.400000e+01 | 0.000000 | 0.000000 | 2.973300e+04 |
| 50% | 3.077600e+04 | 1.520000e+02 | 1.260000e+02 | 0.000000 | 0.000000 | 1.002420e+06 |
| 75% | 4.866800e+04 | 9.640000e+02 | 4.300000e+02 | 0.000000 | 0.000000 | 1.601366e+06 |
| max | 2.746564e+06 | 2.130707e+09 | 8.584298e+09 | 21050.000000 | 20070.000000 | 7.424732e+07 |

8 rows × 92 columns

Fig.6 Overview of Measures Data

So, there is a chance that not all the features are useful. Recursive feature Elimination with Cross Validation (RFECV) gives the actual number of useful features in the Measure's dataset.

*5.1.1 Code for RFECV:*

https://gist.github.com/Tejas-Deo/fb193565ffcaba4e5ee1b5d5a0852e66

The hyperparameters in the code above are chosen after tuning the estimator. The output of RFECV shows that only 63 features out of 94 in the Measures data are useful for the model to predict the output. The top 10 features (features that are the most useful for the model to predict the output) are extracted using Recursive Feature Elimination (RFE) to perform **Feature engineering**. Feature engineering is basically done in order to underline the existing data distribution as this helps the model to learn the data well. By doing the required analysis on these top 10 features, the following 5 features are engineered. They are:

- Sensor12_measure into sensor13_measure
- Sensor12_measure minus sensor13_measure
- Sensor17_measure minus 75th percentile value corresponding to class 0
- Sensor35_measure into sensor17_measure
- Sensor81_measure into sensor82_measure

After checking the Pearson Correlation Coefficient (PCC) values of these engineered features with the target variables, it is evident that no feature is highly correlated (either positively or negatively) with the target variable. But not having a high correlation with target variables indicates that, there is no linear relationship between the engineered features and the target variables. This does not indicate that the engineered features are not useful.

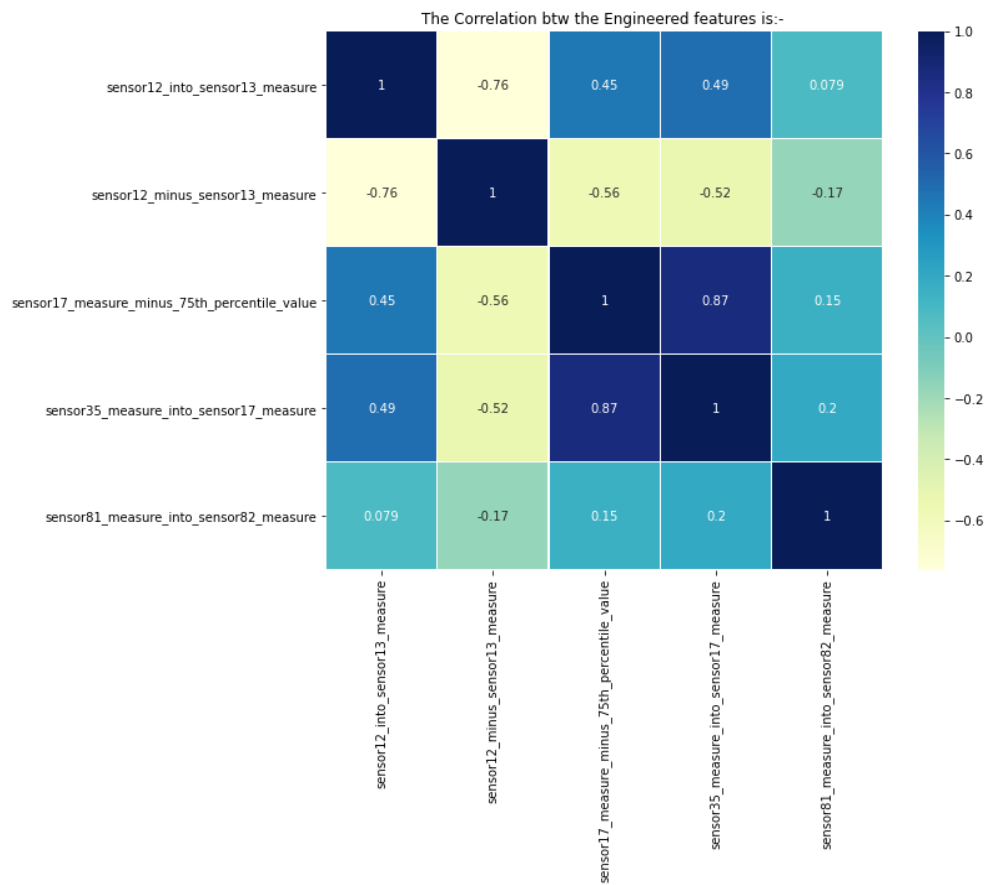

Fig.7 Correlation between the Engineered features

Fig.7 shows that no engineered featured is highly correlated with each other. This is a good sign as the same data distribution is not getting underlined and thus will prevent the model from overfitting. To check the feature importance values of the engineered features, it is imperative to use RFECV on 68 features (previous top 63 features + these 5 engineered features). The output of RFECV gave 29 features which are of utmost importance to the model's prediction. It also includes the following 4 engineered features:

- Sensor12_measure into sensor13_measure
- Sensor12_measure minus sensor13_measure
- Sensor35_measure into sensor17_measure
- Sensor81_measure into sensor82_measure

These top 29 features are responsible for representing the Measures data. And they are further used for data pre-processing and model training. As Measures Data either has values close to 0 or has large values, it is necessary to standardize the data. Now, to split the measures dataset in order to impute the missing values. As there are many NAN values (missing values) in the Measures dataset, the dataset is split into 3 parts based on the percentage of NAN values for data imputation. The Measures data is split into 3 datasets as follows: -

- Dataset D1 has columns having less than 5% NAN values
- Dataset D2 has columns having 5% to 30% of NAN values
- Dataset D3 has columns having 30% to 75% of NAN values

The columns having more than 75% of NAN values were earlier discarded due to insufficient amount of data for imputation. The reason behind choosing this approach is that, this dataset is of the type **MISSING AT RANDOM (MAR)**. Here the values are missing due to some technical error or glitch. Also, the data is coming from different mechanical equipment's and so there is a very high chance that the missing values are CORRELATED with other features in the dataset.

- **Imputation Strategy for D1:** As this dataset has columns having NAN less than 5%, it is not harmful to remove all the rows containing even a single NAN value. This keeps the data in its pure form.
- **Imputation Strategy for D2:** As this dataset has columns having NAN values between 5% to 30%, there is sufficient amount of data for imputation. Iterative Imputer with ExtraTreesRegressor as the estimator is used to impute the data. ExtraTreesRegressor is similar to RandomForestRegressor and the former is much faster. As the estimator is an ensemble model with sufficient amount of data, it can impute the missing values with good precision. The hyper-parameters in the code below are selected after tuning the estimator.

*5.1.2 Code for Iterative Imputer with ExtraTreesRegressor as the estimator:*
https://gist.github.com/Tejas-Deo/6a52cd3e60caec46a7048f3afe4828bc

- **Imputation Strategy for D3:** As this dataset has columns having NAN values between 30% to 75%, there is no sufficient amount of data to use an ensemble model as the estimator in the Iterative Imputer. So, RidgeRegression (Regression with L2 regularization) is used as the estimator to impute the missing values. RidgeRegression will work well as the amount of data that is less and its generalization capability is better as compared to other models. The hyper-parameters in the code below are selected after tuning the estimator.

*5.1.3 Code for Iterative Imputer with RidgeRegression as the estimator:*

https://gist.github.com/Tejas-Deo/91782a5ba6ee70c1852f1f9d9362bf32

As all the Measure's datasets have been imputed, it is necessary to combine all the 3 datasets (D1, D2, D3) on their index values and create a FINAL MEASURES DATASET.

*5.2 Analysis and pre-processing of Histogram Data (shape = (60000, 70))*

There are less than 3% of data points having NAN values in the Histogram dataset. And so, to remove those values to keep the data set pure. Fig.8 shows the output of the "describe()" function to give a brief overview of the Histogram dataset.

| | sensor7_histogram_bin0 | sensor7_histogram_bin1 | sensor7_histogram_bin2 | sensor7_histogram_bin3 | sensor7_histogram_bin4 | sensor7_histogram_bin5 |
|---|---|---|---|---|---|---|
| count | 5.932900e+04 | 5.932900e+04 | 5.932900e+04 | 5.932900e+04 | 5.932900e+04 | 5.932900e+04 |
| mean | 2.216364e+02 | 9.757223e+02 | 8.606015e+03 | 8.859128e+04 | 4.370966e+05 | 1.108374e+06 |
| std | 2.047846e+04 | 3.420053e+04 | 1.503220e+05 | 7.617312e+05 | 2.374282e+06 | 3.262607e+06 |
| min | 0.000000e+00 | 0.000000e+00 | 0.000000e+00 | 0.000000e+00 | 0.000000e+00 | 0.000000e+00 |
| 25% | 0.000000e+00 | 0.000000e+00 | 0.000000e+00 | 0.000000e+00 | 3.080000e+02 | 1.383400e+04 |
| 50% | 0.000000e+00 | 0.000000e+00 | 0.000000e+00 | 0.000000e+00 | 3.672000e+03 | 1.760200e+05 |
| 75% | 0.000000e+00 | 0.000000e+00 | 0.000000e+00 | 0.000000e+00 | 4.952200e+04 | 9.139640e+05 |
| max | 3.376892e+06 | 4.109372e+06 | 1.055286e+07 | 6.340207e+07 | 2.288306e+08 | 1.791880e+08 |

8 rows × 70 columns

Fig.8 Overview of Histogram Data

As the difference between the 75th percentile value and max value is very large, there is **no** way to engineer features from this dataset to underline the existing data distribution. Dropping the histogram data entirely was one option but, to check the usefulness of this data it is essential to tune a model. The data is standardized before passing to the model.

*5.2.1 Code for tuning XGBoost Model on Histogram data:*

https://gist.github.com/Tejas-Deo/81e7312495e4eee285de162f40856876

Hyper-parameter tuning results: n_estimators=900, Macro F1 Score=0.8519

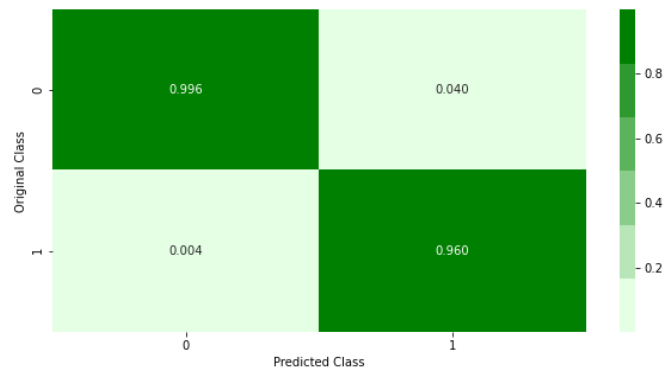

Fig.9 Precision Matrix for Histogram data

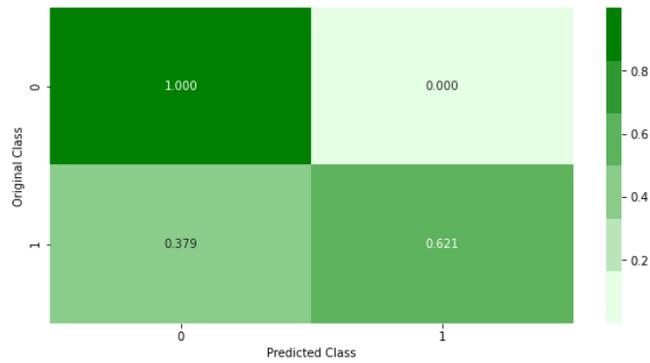

Fig.10 Recall Matrix for Histogram data

From Fig. 10 i.e. Precision and Recall matrix it is evident that this data is certainly useful. Histogram data now will be included in the final dataset alongside Measures data. The only pre-processing that is done on the Histogram data is Standardization. Final data contains both Measures and Histogram data as well. While performing pre-processing on Measures data and Histogram data, it is necessary to store the index values as to combine the datasets at the end to prevent any shape mismatch.

*5.3 First Cut Solution*

The first cut solution consists of using existing model libraries and checking the behaviour/performance of the model. Out of many models tried, this paper will demonstrate the detailed explanation of the best model and the summary for all the remaining models.

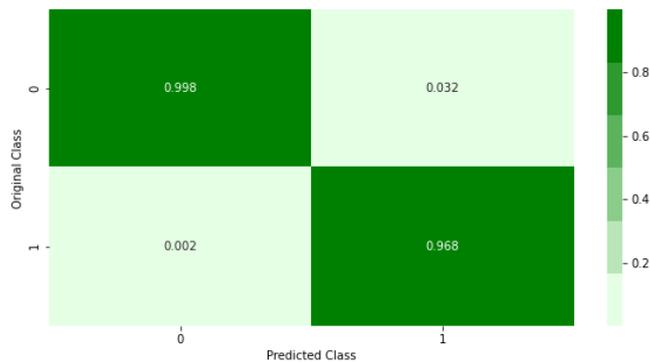

Fig.11 Precision Matrix

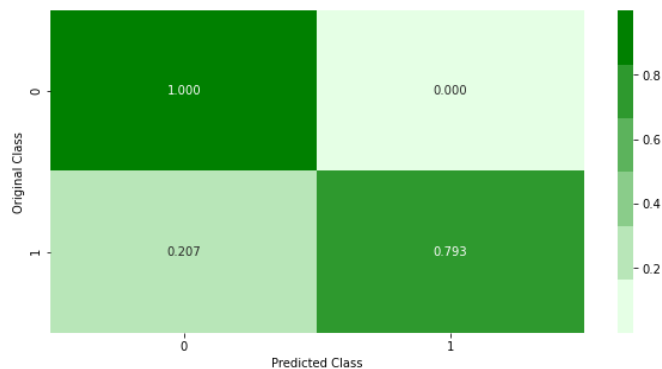

Figure.12 Recall Matrix

As the dataset is highly imbalanced, the macro f1 score will be low even if the model predicts all the data points as class 0 i.e., surface failure, the majority class. The unweighted-XGBoost model gives the highest macro f1 score of 0.89399 and the corresponding hyper-parameters are "n_estimators = 700". From Fig. 12 Macro F1 score, Precision and Recall Matrix it is evident that the model is performing very well on the final dataset and this potentially can be used as a production model.

Other Models tried are:
- XGBoost weighted (giving equal weightage to both the classes)
- RandomForest (un-weighted and weighted)
- CATBoost (un-weighted and weighted)
- AdaBoost
- Training the Model on SMOTE (Synthetic Minority Oversampling technique) data and testing it on the original test data. This method made the model overfit and perform badly on the test data.

*5.4 Custom Model and Architecture Explanation*

To see if the performance can get better on the existing data, this paper also proposes training and testing of Custom Ensemble Models. Just to give an idea, the **BEST MODEL** that the paper proposes among all the models tried (including custom as well as using existing Libraries) is a **Custom Ensemble Model** with 500 Decision Trees as the base estimators and XGBoost with "n_estimators=100" as the Meta Classifier. The architecture for the custom ensemble model (the best model) is as follows: -

*5.4.1 Training Procedure:*

The training procedure is divided into two parts as explained here.

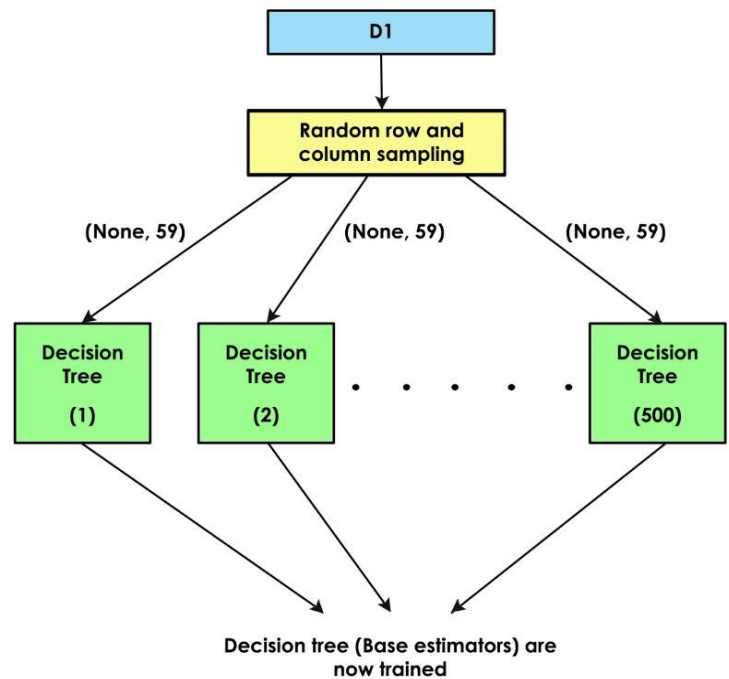

Fig.13 Architecture while Training (PART-1)

*5.4.1.1 Part-1:*

- To split the training data into 2 datasets (D1 and D2) each having equal number of rows and columns.
- On D1, randomly perform row and column sampling with replacement so as to train the 500 Decision Trees which are acting as the base estimators (sampling is done in order to avoid the model from overfitting).
- Then for each decision tree store the corresponding column names as it is required during testing and training the meta classifier.
- Now the base estimators i.e., 500 Decision Trees are trained.

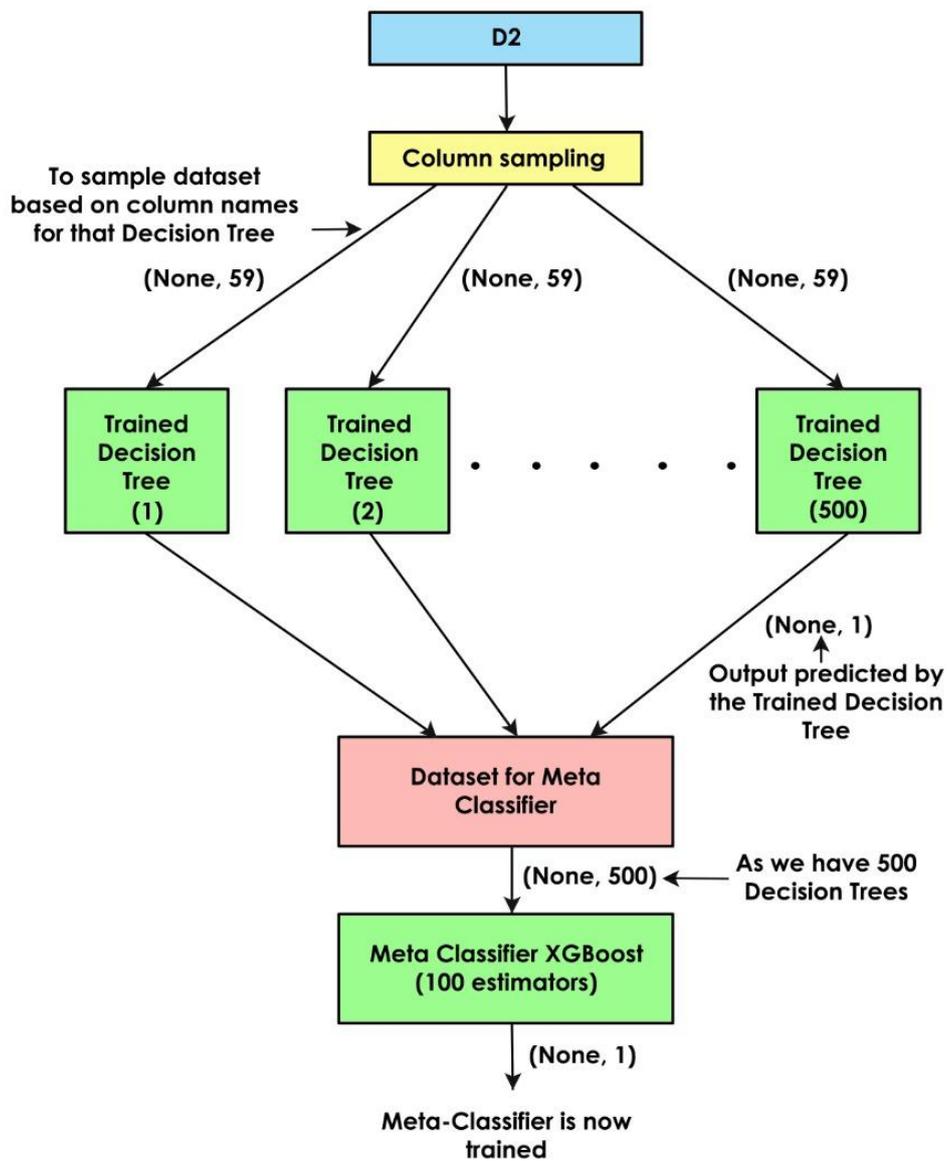

Fig. 14 Architecture while Training (Part-2)

*5.4.1.1 Part-2:*

- Take dataset D2.
- Pass this dataset through the TRAINED BASE ESTIMATORS by performing appropriate column sampling.
- Store the output from each trained base estimator and create a new dataset for the Meta Classifier i.e., XGBoost with "n_estimators = 100".
- Now, use this new dataset to train the Meta Classifier.
- The Meta Classifier is now trained.

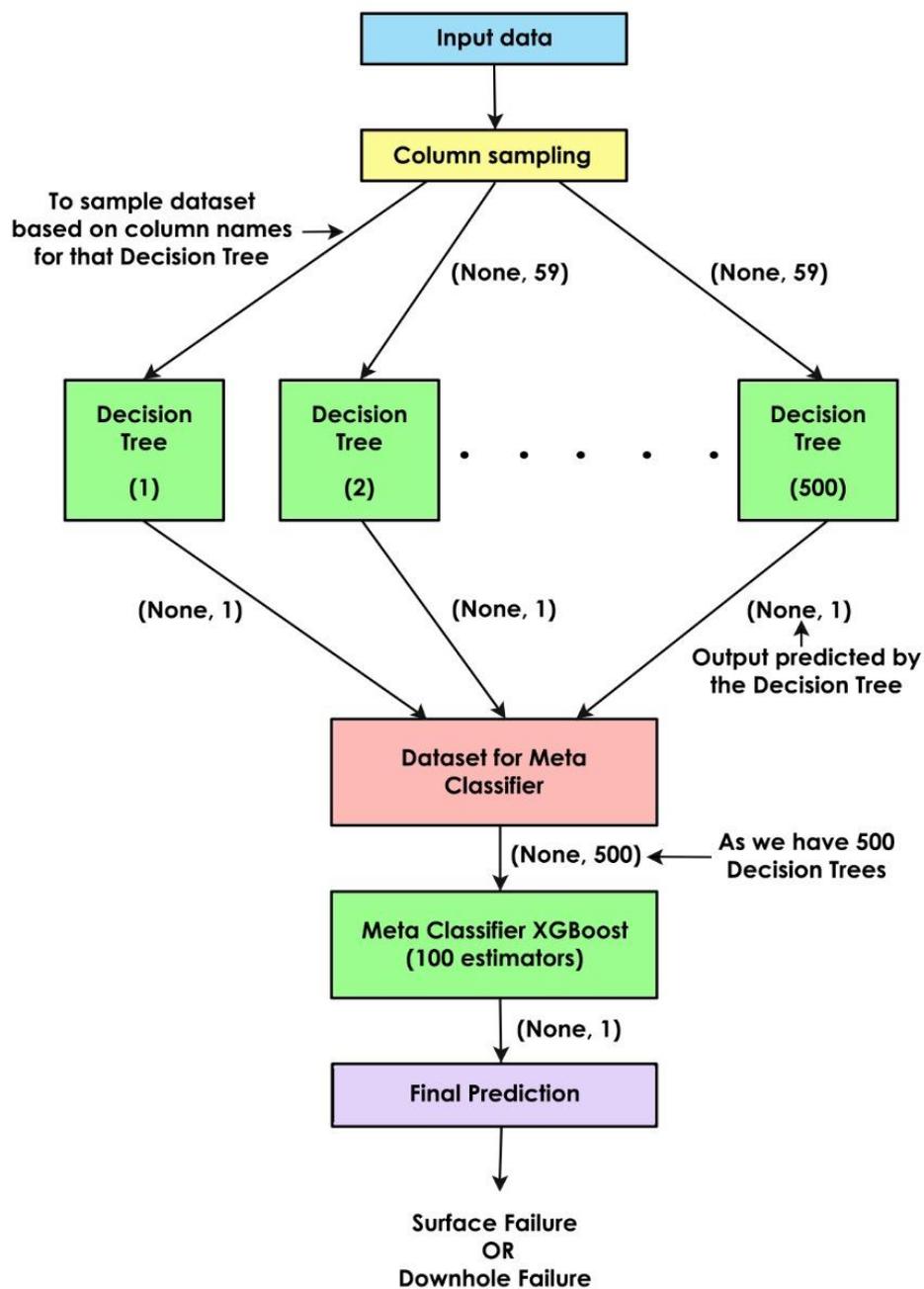

Fig. 15 Architecture while TESTING

*5.4.2 Testing Procedure:*

The testing procedure discussed below is for a single data point, the same can be used for a set of datapoints as well.

- Take a test data point.
- Sample the data point based on column names before passing it to the corresponding trained base estimators.
- Collect the output predicted from each base-estimator and combine them all to make a new data point having shape (1,500).
- Pass this new data point through the trained meta classifier.
- Obtain an output for that data point.

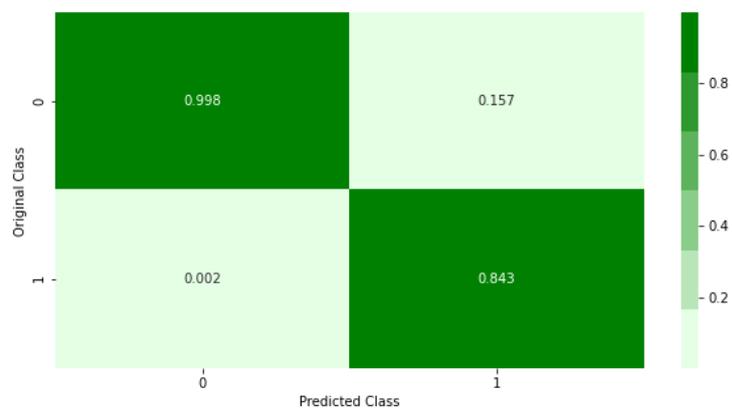

Fig.16 Precision Matrix for Custom Model

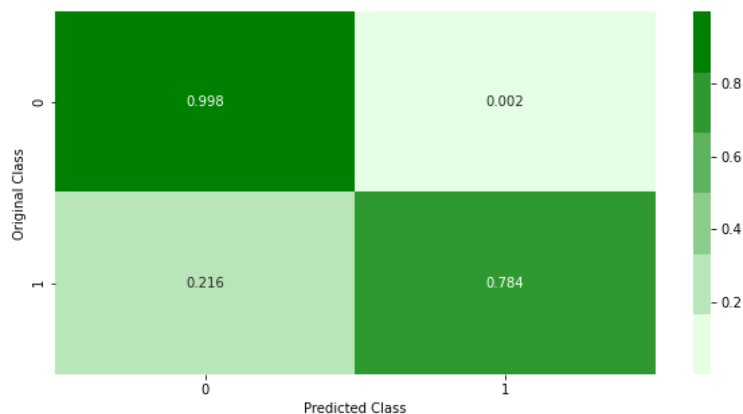

Fig.17 Recall Matrix for Custom Model

The macro f1 score on this custom ensemble model is 0.9053. The number of misclassified points on test data is 0.37443%.

*5.5 Model Comparisons and results*

This section compares the ensemble model's trained and tested using existing model libraries like XGBoost, Scikit-learn, etc and custom ensemble models. Since the size of the dataset is large and there are no low-latency constraints, it is possible to train complex ensemble models.

*5.5.1 Using existing model libraries*

Fig.18 shows the summary of all the models trained and tested using existing model libraries like XGBoost, Scikit-learn, etc. Out of all the models shown in the figure below, unweighted XGBoost model gave the highest macro-f1 score of 0.89399. This performance of unweighted XGBoost model suggests that the macro-f1 score can be further increased by using this model as the meta-classifier. The next subsection mainly discusses the use of unweighted XGBoost model as the meta-classifier.

```
+------------------------------+--------------+----------------+-------------------+-----------------+
|          Model Name          | n_estimators | MACRO F1 Score | Precision Class 1 | Recall Class 1  |
+------------------------------+--------------+----------------+-------------------+-----------------+
|     XGBoost (Un-weighted)    |     700      |    0.89399     |       96.8%       |      79.3%      |
|      XGBoost (weighted)      |     1100     |    0.8874      |        92%        |      79.3%      |
|    CATBoost (un-weighted)    |     400      |    0.87955     |       91.4%       |      73.3%      |
|      CATBoost (weighted)     |     600      |    0.84515     |       84.1%       |      81.9%      |
|           ADABoost           |     500      |    0.87838     |       90.1%       |      70.7%      |
|  RandomForest (un-weighted)  |     200      |    0.8252      |       87.2%       |      70.7%      |
|    RandomForest (weighted)   |     300      |    0.7643      |       73.8%       |      68.1%      |
+------------------------------+--------------+----------------+-------------------+-----------------+
```

Fig.18 Summary of models trained using existing Libraries

*5.5.2 Using Custom Models*

In Fig.19, the "Base Estimator (BE)" column talks about the type of estimators (models) used in the first stage. The number of such models used in the first stage are given in the "No. of Base Estimators" column. The mostly frequently used meta-classifier i.e., model in the second stage is the unweighted XGBoost model as it gave the highest macro-f1 score. Since decision trees can be easily overfitted on a subset of data, they are chosen as the base estimators as the main aim of the base estimators is to learn a subset of data very well. These overfitted models are taken care of in the second stage by the use of meta-classifiers. Also, it is evident from Fig.19 that the use of ensemble models as the base estimators did not work well.

```
+---------------------+---------------------+-----------------------+------------------+----------------+
|  Base Estimator (BE)| Meta Classifier (MC)| No. of Base Estimators| MC Hyperparameter | MACRO F1 Score|
+---------------------+---------------------+-----------------------+------------------+----------------+
|    Decision Tree    |       XGBoost       |          200          |        50        |    0.86599     |
|    Decision Tree    | Logistic Regression |          200          |     C = 0.1      |    0.859       |
|  Random Forest (50) | Logistic Regression |           50          |     C = 100      |    0.8367      |
|  Random Forest (50) |       XGBoost       |           50          |        200       |    0.8442      |
|    Decision Tree    |       XGBoost       |          500          |        100       |    0.90530     |
|    Decision Tree    |  XGBoost (weighted) |          500          |        100       |    0.8348      |
|    Decision Tree    |       XGBoost       |          600          |        100       |    0.89        |
|    Decision Tree    |       XGBoost       |          700          |        100       |    0.8679      |
+---------------------+---------------------+-----------------------+------------------+----------------+
```

Fig.19 Summary of Custom Models trained

The architecture of the Custom ensemble model discussed is the best suitable architecture for the ConocoPhillips Company's data. The specifications for this model are: 500 Decision Trees as the base-estimators, XGBoost with "n_estimators = 100" as the Meta classifier. The macro f1 score on this model is 0.90530, which is higher than any model trained.

**6. Conclusion and Future Work**

Depending upon the percentage of missing data, it is possible to train machine learning models to impute the missing values. Data imputation with standard techniques degrades the quality of data. Since the standard techniques include substitution of missing values with the mean, mode, or median value of that feature. Consequently, dropping all the data points having missing values would decrease the amount of data available and the accuracy of the model. Instead, using a model-based data imputation approach increases the accuracy with which the data is being imputed and makes it more reliable. The use of custom-built models can prove more useful than using standard machine learning libraries depending upon the business constraints. It is always easier to tune the custom-built models according to the changing business requirements. For future work, one can make use of sequential models on the Histogram data like different types of Recurrent neural networks (RNN's) viz LSTM's, Vanilla RNN's, GRU's, etc. Auto-Encoders can also be a good alternative. More in-depth analysis on the Measures data could lead to the engineering of new features.

**Profiles**

1. Link for the GitHub repository: - https://github.com/Tejas-Deo/Predictive-Equipment-Failure
2. Demo video after productionizing the custom ensemble model: - https://www.youtube.com/watch?v=j6eGFZDWw3o
3. LinkedIn profile: - https://www.linkedin.com/in/tejas-deo-799459201/